\documentclass[10pt]{article}
\usepackage[accepted]{rlj} 
\usepackage{amssymb}            
\usepackage{mathtools}          
\usepackage{mathrsfs}           
\usepackage{graphicx}           
\usepackage{subcaption}         
\usepackage[space]{grffile}     
\usepackage{url}                
\usepackage{booktabs}

\title{Prediction-Based Markov Violation Scores for Detecting Non-Markovian Observations in Reinforcement Learning}
\setrunningtitle{Prediction-Based MVS}
\keywords{Markov Property, Conditional Independence Testing, Reinforcement Learning, Non-Markovian Detection.}
\author{Naveen Mysore\textsuperscript{1,2}}
\emails{nmysore@ucsb.edu}
\affiliations{
\textsuperscript{1}Department of Electrical and Computer Engineering, University of California, Santa Barbara, USA\\
\textsuperscript{2}Dyssonance.ai
}

\contribution{
    A prediction-based Markov Violation Score (MVS) using a two-stage pipeline (random forest residualization + ridge comparison) that detects non-Markovian structure in RL trajectories without causal graph construction
}{
    Prior methods rely on causal discovery (e.g., PCMCI) or assume linearity. MVS handles nonlinear dynamics and produces a bounded scalar in $[0,1]$.
}

\contribution{
    Evaluation across 6 environments, 3 algorithms, 6 noise levels, and 10 seeds (960 runs), showing MVS detects violations with Spearman $\rho$ up to $0.78$ and that 13 of 16 pairs exhibit significant reward degradation under noise
}{
    Previous work was limited to few environments and a single algorithm. This is the first large-scale evaluation with significance tests and confidence intervals.
}

\contribution{
    Analysis of an MVS inversion phenomenon where the score decreases as violations increase, characterizing when prediction-based detection fails
}{
    The inversion reveals that flexible first-stage models can absorb violation signals in low-dimensional environments, guiding future diagnostic design.
}

\contribution{
    Utility demonstration: MVS identifies partial observability and guides architecture selection---switching to a history-augmented policy fully recovers lost performance
}{
    Shows MVS provides actionable guidance beyond detection, enabling evidence-based policy architecture decisions.
}

\summary{
A prediction-based Markov Violation Score (MVS) is proposed for detecting non-Markovian observation structure in reinforcement learning. The method works in two stages: a random forest strips out nonlinear Markov-compliant dynamics, then ridge regression checks whether historical observations improve prediction of what remains. MVS is evaluated across six environments (CartPole, Pendulum, Acrobot, HalfCheetah, Hopper, Walker2d), three algorithms (PPO, A2C, SAC), and AR(1) noise at six intensities (16 pairs $\times$ 6 noise levels $\times$ 10 seeds = 960 condition-seed runs). The score successfully detects violations in 7 of 16 environment--algorithm pairs, with the strongest signal in high-dimensional environments (HalfCheetah: $\rho = 0.78$). Phase~2 experiments confirm that violations degrade reward in 13 of 16 conditions. An inversion phenomenon that limits detection in low-dimensional settings is also documented, and its root cause is analyzed as a guide for future work. In a practical utility experiment, MVS is shown to provide actionable guidance for architecture selection: when partial observability renders observations non-Markov, the score identifies the problem and a history-augmented policy chosen accordingly recovers the lost performance.
}

\begin{document}
\makeCover  
\maketitle 
\begin{abstract}
Reinforcement learning algorithms assume that observations satisfy the Markov property, yet real-world sensors frequently violate this assumption through correlated noise, latency, or partial observability. Standard performance metrics conflate Markov breakdowns with other sources of suboptimality, leaving practitioners without tools to detect such violations. This paper introduces a prediction-based Markov Violation Score (MVS) that quantifies non-Markovian structure in observation trajectories. A random forest first removes nonlinear Markov-compliant dynamics; ridge regression then tests whether historical observations reduce prediction error on the residuals beyond what the current observation provides. The resulting score is bounded in $[0,1]$ and requires no causal graph construction. Evaluation spans six environments (CartPole, Pendulum, Acrobot, HalfCheetah, Hopper, Walker2d), three algorithms (PPO, A2C, SAC), controlled AR(1) noise at six intensity levels, and 10 seeds per condition. In post-hoc detection, 7 of 16 environment--algorithm pairs---primarily high-dimensional locomotion tasks---show significant positive monotonicity between noise intensity and MVS (Spearman $\rho$ up to $0.78$, confirmed under repeated-measures analysis); under training-time noise, 13 of 16 pairs exhibit statistically significant reward degradation. An inversion phenomenon is documented in low-dimensional environments where the random forest absorbs the noise signal, causing MVS to decrease as true violations grow---a failure mode analyzed in detail. A practical utility experiment demonstrates that MVS correctly identifies partial observability and guides architecture selection, fully recovering performance lost to non-Markovian observations. Source code to reproduce all results is available at \url{https://github.com/NAVEENMN/Markovianes}.
\end{abstract}

\section{Introduction}
\label{sec:introduction}

Reinforcement learning (RL) algorithms overwhelmingly assume that observations satisfy the Markov property: the current observation, together with the current action, is sufficient to predict the distribution over next observations and rewards~\citep{sutton_reinforcement_1998}. This assumption underpins convergence guarantees, the policy gradient theorem, and the design of virtually all model-free and model-based methods. In practice, though, real-world observations are routinely corrupted by sensor noise, communication delays, and partial observability~\citep{wisniewski_benchmarking_2024}, any of which can introduce temporal correlations that break the Markov property.

When that happens, value functions conditioned on the current state alone become systematically biased, and policy gradients may point in wrong directions. The trouble is that standard evaluation metrics---episodic return, sample efficiency, convergence rate---cannot tell a practitioner \emph{why} an agent is underperforming. An agent struggling because its observations are non-Markovian looks identical, by these metrics, to one struggling with reward sparsity or function approximation error. Despite the practical importance of this failure mode, the RL community lacks standard diagnostic tools for detecting Markov violations in observation trajectories.

This paper proposes a \emph{prediction-based Markov Violation Score} (MVS) to address this gap. The core idea is straightforward: first, a random forest captures whatever structure is predictable from the current state--action pair alone; then, ridge regression checks whether lagged observations carry additional predictive signal about the residuals. If they do, history matters and the Markov property is violated. The resulting score lies in $[0,1]$ and requires no causal graph construction.

Three main contributions are made. First, the two-stage MVS pipeline is developed and evaluated across six Gymnasium environments (CartPole, Pendulum, Acrobot, HalfCheetah, Hopper, Walker2d), three algorithms (PPO, A2C, SAC), six AR(1) noise intensities, and 10 seeds per condition---960 runs in total (Sections~\ref{sec:mvs} and~\ref{sec:experiments}). Second, in the 7 environment--algorithm pairs where MVS correctly tracks violations (Spearman $\rho$ up to $0.78$), higher MVS corresponds to worse policy performance, and 13 of 16 pairs show statistically significant reward degradation under noise. However, an inversion phenomenon is also observed in low-dimensional environments where the random forest absorbs the noise signal rather than leaving it for the second stage (Section~\ref{subsec:inversion}). This inversion was initially surprising, and its root cause is analyzed in detail, since understanding when the method fails is as important as demonstrating when it works. Third, MVS is shown to be actionable: in a controlled partial-observability experiment on CartPole, an MVS-guided strategy that switches to a history-augmented policy when violations are detected fully recovers the lost performance (Section~\ref{subsec:utility}).

The primary focus is \emph{diagnosis}---giving practitioners a tool to answer ``are my observations non-Markovian?''---rather than a complete remedy. That said, the utility experiment demonstrates that even a simple threshold on MVS can drive architecture decisions with substantial impact on reward.

\textbf{Paper organization.} Section~\ref{sec:related-works} surveys related work. Section~\ref{sec:background} introduces the Markov property and its connection to conditional independence. Section~\ref{sec:mvs} defines the prediction-based MVS. Section~\ref{sec:experiments} presents experiments, including the utility demonstration. Section~\ref{sec:limitations} discusses limitations, and Section~\ref{sec:conclusion} concludes.

\section{Related Works}
\label{sec:related-works}

\paragraph{Robust RL and observation noise.} Real-world RL deployments frequently encounter noisy or corrupted observations. The robust RL literature addresses this through adversarial training~\citep{pinto_robust_2017}, model-based methods for incomplete or noisy observations~\citep{wang_robust_2019}, and distributionally robust formulations~\citep{panaganti_robust_2022, liu_distributionally_2022}. Separately, the effect of correlated action noise on exploration and performance has been studied in continuous control~\citep{hollenstein_colored_2024, hollenstein_action_nodate}, and selective noise injection has been proposed as a regularizer to improve generalization~\citep{igl_generalization_2019}. These lines of work all evaluate robustness or exploration quality through downstream task performance; none of them tell a practitioner \emph{whether} or \emph{how severely} the Markov property has been violated.

\paragraph{Partial observability and POMDPs.} When the full state is not directly observable, the problem becomes a POMDP~\citep{lauri_partially_2023}. Solutions range from belief-state methods to identifying tractable POMDP subclasses with provable sample efficiency~\citep{liu_when_2022}. These methods can mitigate non-Markovian structure by maintaining history or exploiting structural assumptions, but they do not quantify how much history dependence is present in the first place. The present work is complementary: MVS diagnoses the violation; POMDP methods address it.

\paragraph{Conditional independence testing and Granger causality.} The statistics literature provides many tools for testing conditional independence, from kernel-based tests~\citep{zhang_kernel-based_2011} to classifier-based approaches. Granger causality~\citep{granger_investigating_1969} asks whether past values of one series improve prediction of another---closely related to the approach taken here. Nonlinear extensions via neural networks~\citep{tank_neural_2022} or random forests handle richer dynamics than linear methods. MVS builds on this tradition but targets the RL setting specifically, where the question reduces to whether \emph{any} historical observation helps predict the next state beyond what the current state already provides.

\paragraph{Markov property testing in RL.} Directly testing the Markov property in RL trajectories has received limited attention. \citet{shi_does_2020} proposed a Forward-Backward Learning procedure that tests the Markov assumption in sequential decision making without parametric assumptions on the joint distribution, with theoretical validity guarantees. However, that test is designed for binary hypothesis testing (Markov or not) and does not produce a graded severity score. Constraint-based causal discovery methods such as PCMCI~\citep{runge_discovering_2022} can detect multi-lag dependencies; the framework supports both linear and nonlinear conditional independence tests, though the default partial-correlation test assumes linearity. \citet{mysore_markov_2025} applied PCMCI with partial correlation to quantify first-order Markov violations in noisy RL, demonstrating that such diagnostics are useful but inheriting the linearity constraints of that specific test. The prediction-based approach proposed here sidesteps these issues: the random forest first stage is nonparametric, and history dependence is tested directly through prediction error comparison rather than graph construction.

\paragraph{Positioning.} Robust RL assumes violations exist and builds defenses; causal discovery infers graphical structure. MVS sits between the two, providing a single scalar answer to a more focused question: does the current observation suffice, or does history help? For an RL practitioner, that question is often more actionable than a full causal graph, and the prediction-based formulation handles nonlinear dynamics naturally.

\section{Background}
\label{sec:background}

\subsection{Markov Property and Markov Decision Processes}
\label{subsec:markov-mdp}

A discrete-time stochastic process $\{X_t\}_{t=0}^{\infty}$ satisfies the \emph{Markov property} if the future state $X_{t+1}$ is conditionally independent of all prior states given the current state:
\[
    P\bigl(X_{t+1} \mid X_t, X_{t-1}, \ldots, X_0\bigr)
    \;=\;
    P\bigl(X_{t+1} \mid X_t\bigr).
\]
In RL, this applies to the state variable $S_t$. When the environment is Markov,
\begin{multline*}
    P\bigl(S_{t+1} = s', R_{t+1} = r \,\mid\, S_t = s, A_t = a, \ldots, S_0, A_0\bigr) \\
    = P\bigl(S_{t+1} = s', R_{t+1} = r \,\mid\, S_t = s, A_t = a\bigr),
\end{multline*}
so only the current state $S_t$ and action $A_t$ determine what happens next.

When noise or partial observability corrupts $S_t$, the observed signal $O_t$ may no longer carry enough information, and higher-order dependencies can emerge in the observation stream---even though the underlying state dynamics remain Markov. Detecting such observation-level dependencies is the central goal of this work. Throughout, ``Markov violation'' refers to non-Markovian structure in observations, not a breakdown of the latent state dynamics.

\subsection{Conditional Independence as a Test for Markov Structure}
\label{subsec:ci-markov}

Two random variables $X$ and $Y$ are \emph{conditionally independent} given $Z$ if $P(X \mid Y, Z) = P(X \mid Z)$. The Markov property is exactly such a statement: $S_{t+1} \perp\!\!\!\perp \{S_0, \ldots, S_{t-1}\} \mid S_t$. Whenever knowledge of past states improves prediction of the next state beyond what the current state provides, this independence is violated.

This observation suggests a practical detection strategy. Rather than constructing a causal graph or computing partial correlations, one can directly test whether historical observations carry predictive information that the current observation misses. The next section formalizes this idea.

\section{Prediction-Based Markov Violation Score}
\label{sec:mvs}

This section describes the Markov Violation Score (MVS), a scalar that quantifies how much an observation trajectory departs from Markov. The method has two stages: first strip out whatever the current state can predict, then check whether past observations help explain what remains.

\subsection{Stage 1: Nonlinear Markov Removal}
\label{subsec:stage1}

Given a trajectory of observations $\{o_1, o_2, \ldots, o_T\}$ and actions $\{a_1, a_2, \ldots, a_T\}$, two sets of features are constructed. The \emph{Markov features} $\mathbf{x}_t^{(M)} = [o_t, a_t]$ contain only the information that would suffice if the process were Markov. The \emph{history features} $\mathbf{x}_t^{(H)} = [o_t, a_t, o_{t-1}, a_{t-1}, \ldots, o_{t-k+1}, a_{t-k+1}]$ additionally include $k-1$ lagged observation--action pairs. The target is always the next observation $y_t = o_{t+1}$.

In Stage~1, a random forest regressor predicts $y_t$ from the Markov features alone:
\[
    \hat{y}_t^{\mathrm{RF}} = f_{\mathrm{RF}}\bigl(\mathbf{x}_t^{(M)}\bigr).
\]
The forest uses 200 trees with max depth 10 and a minimum of 5 samples per leaf. To avoid information leakage, residuals are computed using out-of-bag (OOB) predictions:
\[
    r_t = y_t - \hat{y}_t^{\mathrm{OOB}}.
\]
After this stage, the residuals $\{r_t\}$ have been stripped of everything predictable from the current observation and action. Whatever predictable structure remains must come from history---exactly the signal that indicates a Markov violation.

\subsection{Stage 2: Ridge Comparison}
\label{subsec:stage2}

The residuals are split into training (first 70\%) and test (last 30\%) sets chronologically, and two ridge regressions are fit:

\paragraph{Markov ridge model.} $g_M$ predicts residuals from Markov features:
\[
    \hat{r}_t^{(M)} = g_M\bigl(\mathbf{x}_t^{(M)}\bigr).
\]
This baseline picks up any linear Markov structure that the random forest may have missed.

\paragraph{History ridge model.} $g_H$ predicts residuals from history features:
\[
    \hat{r}_t^{(H)} = g_H\bigl(\mathbf{x}_t^{(H)}\bigr).
\]
If the process is truly Markov, $g_H$ should do no better than $g_M$.

Both models use RidgeCV with 20 regularization values $\alpha \in \{10^{-2}, 10^{-1.6}, \ldots, 10^{5}\}$ selected via leave-one-out cross-validation (LOO). LOO on sequential data can leak temporal information through shared lagged features; blocked or rolling cross-validation would provide a stricter guard against this, at the cost of less efficient use of the training set. Test-set errors are:
\[
    \mathrm{MSE}_M = \frac{1}{n_{\mathrm{test}}} \sum_{t \in \mathcal{T}_{\mathrm{test}}} \bigl(r_t - \hat{r}_t^{(M)}\bigr)^2, \qquad
    \mathrm{MSE}_H = \frac{1}{n_{\mathrm{test}}} \sum_{t \in \mathcal{T}_{\mathrm{test}}} \bigl(r_t - \hat{r}_t^{(H)}\bigr)^2.
\]

\subsection{MVS Definition}
\label{subsec:mvs-def}

The Markov Violation Score is the fractional reduction in prediction error from adding history:
\begin{equation}
\label{eq:mvs}
    \mathrm{MVS} = \mathrm{clip}\!\left(\frac{\mathrm{MSE}_M - \mathrm{MSE}_H}{\mathrm{MSE}_M},\; 0,\; 1\right).
\end{equation}

A few properties are worth noting. The score is bounded in $[0,1]$ by construction: negative values (where the history model performs worse, typically due to estimation noise in finite samples) are clipped to zero. When the process is Markov, the history model gains nothing and $\mathrm{MVS} \approx 0$. Intuitively, MVS measures the relative reduction in test-set prediction error from adding historical observations. The random forest first stage is important because it handles nonlinear-but-Markov dynamics (e.g., the trigonometric relationships in Pendulum) that would otherwise produce false positives. The ridge second stage, with cross-validated regularization, guards against overfitting to spurious history correlations in finite samples.

\paragraph{Design choices.} A history depth of $k = 3$ (two additional lags beyond the current) is used, along with a 70/30 train--test split and per-dimension MVS computation averaged across observation dimensions. AR(1) noise induces dependence at all lags (decaying as $\alpha^\ell$); $k = 3$ captures the strongest portion (lags 1--2) while keeping computation manageable.

\subsection{AR(1) Noise as a Controlled Markov Violation}
\label{subsec:ar-noise}

To evaluate MVS under violations of known severity, autoregressive noise is injected into observations. At each time step, each dimension $i$ is corrupted:
\[
    \tilde{o}_t^{(i)} = o_t^{(i)} + z_t^{(i)}, \qquad z_t^{(i)} = \alpha \cdot z_{t-1}^{(i)} + \epsilon_t^{(i)}, \quad \epsilon_t^{(i)} \sim \mathcal{N}(0, 1),
\]
where $\alpha \in [0, 1)$ controls the autocorrelation. When $\alpha = 0$, the noise is i.i.d.\ and introduces no temporal correlation into the observation stream. (Strictly, even i.i.d.\ additive noise can make observations non-Markov in the hidden-Markov-model sense, since $\tilde{o}_t$ is a noisy function of the latent state; however, no \emph{additional} history dependence is created by the noise process itself, and MVS is empirically near zero in this condition.) For $\alpha > 0$, the noise process $z_t$ carries information from previous time steps into $\tilde{o}_t$, introducing temporal dependence that grows with $\alpha$.

An important distinction: the underlying environment dynamics remain Markov in the augmented state $(S_t, z_t)$. However, because the noise process $z_t$ is hidden from the agent, the \emph{observation stream} $\{\tilde{o}_t\}$ is non-Markov---the current corrupted observation alone does not determine the conditional distribution of the next. MVS targets exactly this observation-level non-Markovity, which is what a practitioner encounters when working with sensor data.

This setup gives two experimental phases:
\begin{enumerate}
    \item \textbf{Phase~1 (Post-hoc detection):} Policies are trained on clean observations. AR(1) noise is injected into collected trajectories afterward, and MVS is computed. This isolates detection capability from any policy adaptation effects.
    \item \textbf{Phase~2 (Training under noise):} Policies are trained from scratch with AR(1) noise present throughout, measuring how Markov violations affect learning.
\end{enumerate}

\section{Experiments and Results}
\label{sec:experiments}

MVS is evaluated across six RL environments, three algorithms, and six noise intensities. Phase~1 asks whether MVS can detect controlled Markov violations in post-hoc trajectories; Phase~2 measures how those same violations affect policy learning.

\subsection{Experimental Setup}
\label{subsec:setup}

\paragraph{Environments.} Table~\ref{tab:environments} summarizes the six environments, which span classic control and continuous locomotion from OpenAI Gymnasium~\citep{towers_gymnasium_2024}.

\begin{table}[htbp]
    \centering
    \small
    \caption{Environment summary. Observation dimensionality ranges from 3 (Pendulum) to 17 (HalfCheetah, Walker2d). SAC is used only for continuous-action environments.}
    \label{tab:environments}
    \begin{tabular}{lcccc}
        \toprule
        \textbf{Environment} & \textbf{Obs.\ Dim} & \textbf{Action Space} & \textbf{Algorithms} & \textbf{Training Steps} \\
        \midrule
        CartPole-v1     & 4  & Discrete (2)    & PPO, A2C       & 50k \\
        Pendulum-v1     & 3  & Continuous (1)   & PPO, A2C, SAC  & 450k \\
        Acrobot-v1      & 6  & Discrete (3)    & PPO, A2C       & 50k \\
        HalfCheetah-v4  & 17 & Continuous (6)   & PPO, A2C, SAC  & 1M \\
        Hopper-v4       & 11 & Continuous (3)   & PPO, A2C, SAC  & 1M \\
        Walker2d-v4     & 17 & Continuous (6)   & PPO, A2C, SAC  & 1M \\
        \bottomrule
    \end{tabular}
\end{table}

\paragraph{Algorithms.} PPO~\citep{schulman_proximal_2017} and A2C are used across all six environments, with SAC~\citep{haarnoja_soft_2018} additionally applied to the four continuous-action environments. SAC is excluded from CartPole and Acrobot because the \texttt{stable-baselines3}~\citep{raffin_stable-baselines3_2021} SAC implementation requires continuous action spaces. All agents use two-hidden-layer MLP policies, giving 16 environment--algorithm pairs in total.

\paragraph{Noise protocol.} AR(1) noise (Section~\ref{subsec:ar-noise}) is applied to all observation dimensions at six autocorrelation levels: $\alpha \in \{0.0, 0.1, 0.3, 0.5, 0.7, 0.9\}$. The $\alpha = 0$ condition is the clean baseline.

\paragraph{Seeds and evaluation.} Every condition is run with 10 independent seeds. Means with 95\% confidence intervals are reported, and significance is assessed via Spearman rank correlation (Phase~1) and Welch's $t$-test (Phase~2) at the $p < 0.05$ level. Because multiple environment--algorithm pairs are tested (16 in Phase~1, 15 in Phase~2), Benjamini--Hochberg false discovery rate (FDR) correction is applied; significance counts reported throughout refer to FDR-adjusted $q$-values.


\subsection{Phase 1: MVS Sensitivity to Noise Intensity}
\label{subsec:phase1}

Policies are first trained on clean observations. After training, trajectories are collected, AR(1) noise is injected post-hoc at each $\alpha$ level, and MVS is computed. This design isolates MVS detection from any confounding policy adaptation.

\begin{figure}[htbp]
    \centering
    \includegraphics[width=\textwidth]{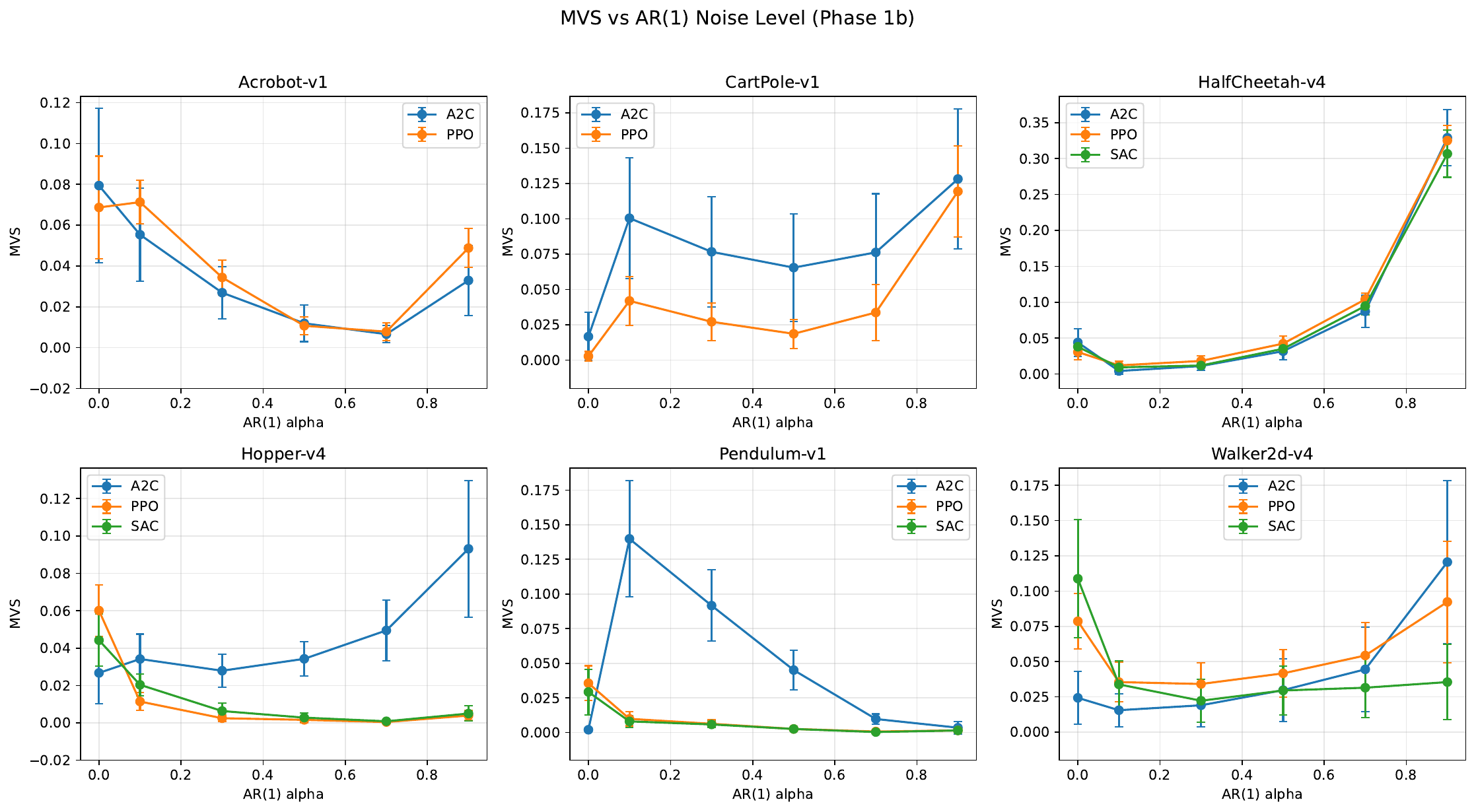}
    \caption{\textbf{Phase 1: MVS vs.\ noise intensity.} Each panel shows one environment; lines represent different algorithms. Error bars are 95\% CIs over 10 seeds. MVS increases monotonically with $\alpha$ in HalfCheetah and CartPole. In Pendulum, Hopper (PPO/SAC), and Acrobot it \emph{decreases}---the inversion phenomenon discussed in Section~\ref{subsec:inversion}.}
    \label{fig:phase1}
\end{figure}

Figure~\ref{fig:phase1} plots MVS against $\alpha$ for all 16 pairs. Table~\ref{tab:monotonicity} gives the Spearman correlations.

\begin{table}[htbp]
    \centering
    \small
    \caption{\textbf{Phase 1 monotonicity.} Spearman $\rho$ between noise intensity $\alpha$ and MVS, pooled across seeds ($n = 60$ per pair). Seven pairs show the expected positive trend (HalfCheetah strongest at $\rho = 0.78$). Eight show significant inversion. See text for repeated-measures analysis that accounts for within-seed dependence.}
    \label{tab:monotonicity}
    \begin{tabular}{llrrl}
        \toprule
        \textbf{Environment} & \textbf{Algorithm} & \textbf{Spearman $\rho$} & \textbf{$p$-value} & \textbf{Direction} \\
        \midrule
        \multicolumn{5}{l}{\emph{Positive monotonicity (7 pairs):}} \\
        HalfCheetah-v4 & PPO & 0.776 & $<10^{-12}$ & \checkmark \\
        HalfCheetah-v4 & SAC & 0.700 & $<10^{-9}$  & \checkmark \\
        HalfCheetah-v4 & A2C & 0.664 & $<10^{-8}$  & \checkmark \\
        CartPole-v1    & PPO & 0.554 & $<10^{-5}$  & \checkmark \\
        Hopper-v4      & A2C & 0.520 & $<10^{-4}$  & \checkmark \\
        Walker2d-v4    & A2C & 0.507 & $<10^{-4}$  & \checkmark \\
        CartPole-v1    & A2C & 0.447 & $<10^{-3}$  & \checkmark \\
        \midrule
        \multicolumn{5}{l}{\emph{Inverted (8 pairs):}} \\
        Pendulum-v1    & PPO & $-0.760$ & $<10^{-12}$ & $\times$ \\
        Pendulum-v1    & SAC & $-0.756$ & $<10^{-12}$ & $\times$ \\
        Hopper-v4      & SAC & $-0.724$ & $<10^{-10}$ & $\times$ \\
        Hopper-v4      & PPO & $-0.643$ & $<10^{-7}$  & $\times$ \\
        Acrobot-v1     & PPO & $-0.497$ & $<10^{-4}$  & $\times$ \\
        Acrobot-v1     & A2C & $-0.466$ & $<10^{-3}$  & $\times$ \\
        Walker2d-v4    & SAC & $-0.330$ & $0.010$     & $\times$ \\
        Pendulum-v1    & A2C & $-0.280$ & $0.030$     & $\times$ \\
        \midrule
        \multicolumn{5}{l}{\emph{Not significant (1 pair):}} \\
        Walker2d-v4    & PPO & 0.070 & $0.596$     & --- \\
        \bottomrule
    \end{tabular}
\end{table}

\paragraph{Repeated-measures validation.} Because Phase~1 injects six noise levels into the same trajectory per seed, the 60 points per pair are not independent. To verify that the pooled $p$-values are not artifacts, Spearman $\rho$ was computed within each seed (across the 6 $\alpha$ values). For all 7 positive pairs, all 10 seeds show positive $\rho$ (sign-test $p = 0.002$). Page's $L$ trend test---which directly tests ordered alternatives under repeated measures---confirms significance for all 7 ($p < 10^{-4}$), with within-seed median $\rho$ from $0.54$ to $0.83$.

\paragraph{Specificity check.} MVS was also computed on trajectories from random (untrained) policies under the same noise. Random-policy MVS is near zero everywhere, confirming that the score does not fire on unstructured trajectories. The nonzero scores seen in trained-policy runs reflect a genuine interaction between the AR(1) noise and the structure that the learned policy imposes on trajectories. A stronger specificity test---Markov but nonlinear dynamics with i.i.d.\ noise under trained policies---is left to future work.


\subsection{Inverted MVS: When Detection Fails}
\label{subsec:inversion}

In eight of 16 pairs, MVS moves in the \emph{wrong} direction: it \emph{decreases} as $\alpha$ increases (Table~\ref{tab:monotonicity}). This was initially surprising and prompted a detailed investigation.

\paragraph{Root cause.} The problem lies in Stage~1. In low-dimensional environments with highly regular trained-policy trajectories, the random forest is flexible enough to fit not just the true Markov dynamics but also the AR(1) noise pattern riding on top of them. Once the RF captures that noise structure, the residuals are scrubbed of the very signal Stage~2 needs. As $\alpha$ grows and the noise becomes a larger fraction of the total signal, the RF fits it more aggressively---hence the inverted relationship.

\paragraph{When it happens.} Two factors predict inversion: (1)~low observation dimensionality, which gives the RF fewer features and makes noise patterns easier to memorize, and (2)~highly structured clean-policy trajectories, which provide a regular backdrop against which AR(1) noise stands out. HalfCheetah, with 17 dimensions and noisier dynamics, is largely immune.

\paragraph{Implications.} This is a fundamental limitation of any two-stage design where the first stage is flexible enough to absorb the violation signal. Potential mitigations---restricting RF capacity, using a linear first stage in low-dimensional settings, ensemble strategies---are discussed in Section~\ref{sec:limitations}.


\subsection{Phase 2: Impact of Markov Violations on Reward}
\label{subsec:phase2}

Phase~2 asks whether the violations MVS is designed to detect actually matter for learning. Agents are trained from scratch with AR(1) noise present throughout. Clean ($\alpha = 0$) versus heavily noised ($\alpha = 0.9$) final reward is compared using Welch's $t$-test across 10 seeds.

\begin{figure}[htbp]
    \centering
    \includegraphics[width=\textwidth]{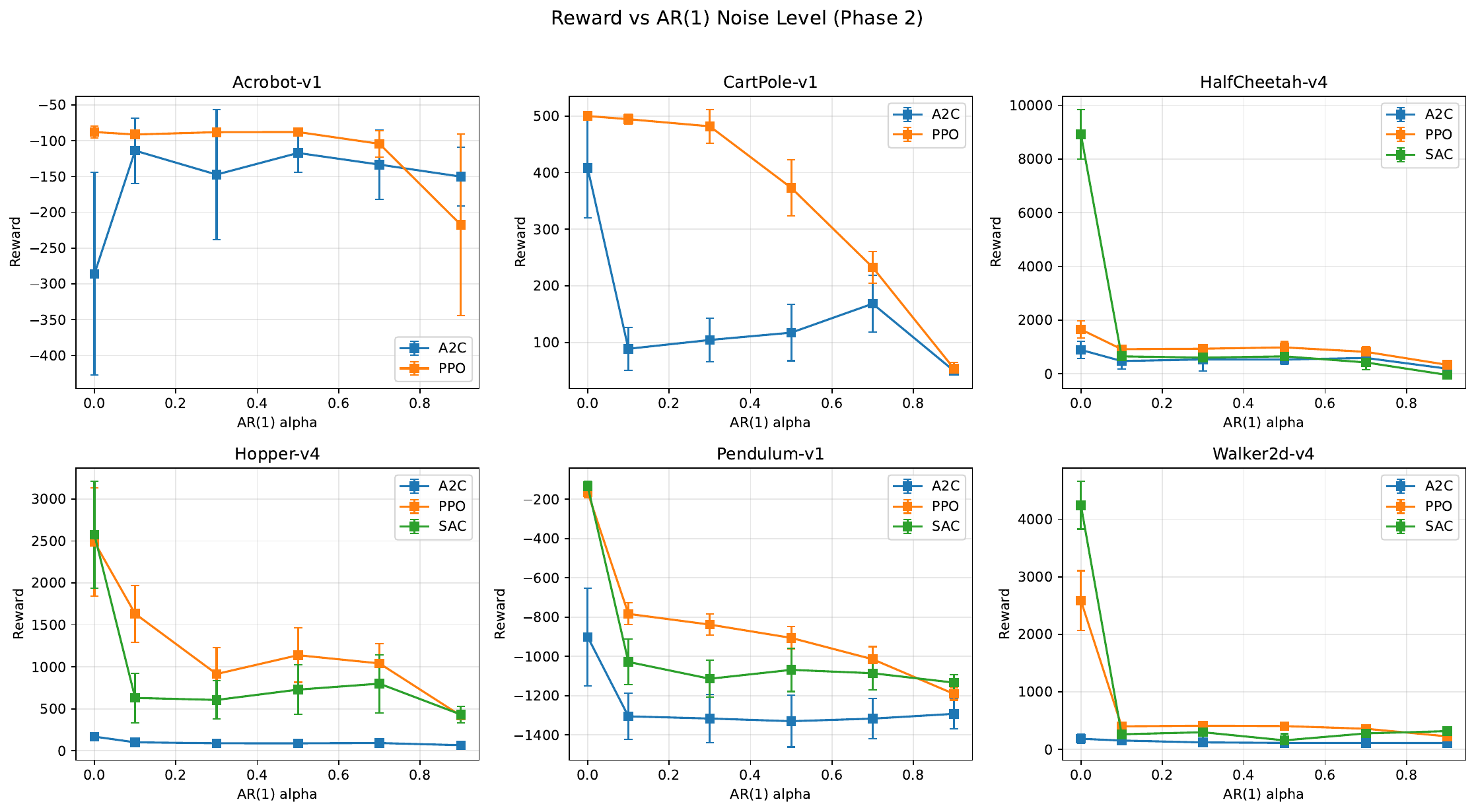}
    \caption{\textbf{Phase 2: Reward vs.\ noise intensity.} AR(1) noise during training degrades final performance across nearly all conditions. The worst collapses: HalfCheetah-SAC drops from 8920 to $-42$; Walker2d-SAC from 4244 to 318; CartPole-PPO from 500 to 53.}
    \label{fig:phase2}
\end{figure}

Figure~\ref{fig:phase2} plots reward against $\alpha$; Table~\ref{tab:welch} gives the statistical comparisons.

\begin{table}[htbp]
    \centering
    \small
    \caption{\textbf{Phase 2: Welch's $t$-test.} Clean ($\alpha=0$) vs.\ noised ($\alpha=0.9$) reward, 10 seeds. Thirteen of 16 pairs show significant degradation after Benjamini--Hochberg FDR correction ($q < 0.05$). Rewards rounded to integers.}
    \label{tab:welch}
    \begin{tabular}{llrrrl}
        \toprule
        \textbf{Environment} & \textbf{Algorithm} & \textbf{Clean Reward} & \textbf{Noised Reward} & \textbf{$t$-stat} & \textbf{Significant} \\
        \midrule
        CartPole-v1    & PPO & 500    & 53    & 94.3  & Yes \\
        Pendulum-v1    & PPO & $-166$   & $-1191$  & 57.4  & Yes \\
        Pendulum-v1    & SAC & $-133$   & $-1134$  & 48.2  & Yes \\
        HalfCheetah-v4 & SAC & 8920   & $-42$    & 22.1  & Yes \\
        Walker2d-v4    & SAC & 4244   & 318   & 21.3  & Yes \\
        Walker2d-v4    & PPO & 2587   & 227   & 10.2  & Yes \\
        HalfCheetah-v4 & PPO & 1654   & 336   & 9.25  & Yes \\
        CartPole-v1    & A2C & 408    & 50    & 9.24  & Yes \\
        Hopper-v4      & SAC & 2574   & 432   & 7.53  & Yes \\
        Hopper-v4      & PPO & 2487   & 418   & 7.24  & Yes \\
        Hopper-v4      & A2C & 169    & 67    & 5.22  & Yes \\
        HalfCheetah-v4 & A2C & 893    & 193   & 4.70  & Yes \\
        Pendulum-v1    & A2C & $-901$   & $-1292$  & 3.41  & Yes \\
        \midrule
        Acrobot-v1     & PPO & $-88$    & $-217$   & 2.31  & No ($q=0.053$) \\
        Walker2d-v4    & A2C & 186    & 112   & 1.95  & No ($q=0.08$) \\
        Acrobot-v1     & A2C & $-286$   & $-150$   & $-2.09$ & No ($q=0.07$) \\
        \bottomrule
    \end{tabular}
\end{table}

The damage is substantial. The worst-hit condition, HalfCheetah-SAC, collapses from 8920 to $-42$---essentially complete failure. Walker2d-SAC drops 93\% (4244 to 318), CartPole-PPO 89\% (500 to 53). The three non-significant pairs after FDR correction are all borderline ($q < 0.08$) and involve either A2C in environments where it already performs modestly, or Acrobot where the absolute reward scale is small.

A caveat: AR(1) noise with larger $\alpha$ has higher marginal variance, so part of the degradation may reflect noise power rather than temporal correlation per se. Matching marginal variance across $\alpha$ while varying only autocorrelation would isolate the non-Markov contribution; this more controlled design is left to future work.


\subsection{Combined Analysis}
\label{subsec:combined}

\begin{figure}[htbp]
    \centering
    \includegraphics[width=0.7\textwidth]{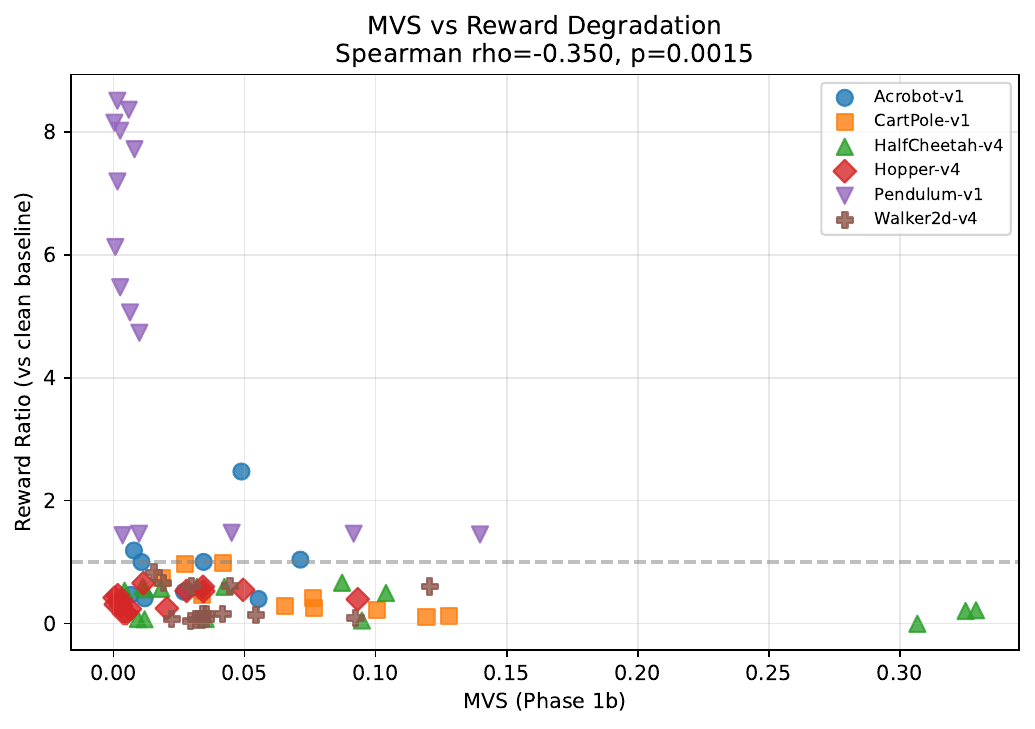}
    \caption{\textbf{Combined: MVS vs.\ reward ratio.} Each point is one environment--algorithm--noise-level condition. In environments where MVS correctly tracks violations (e.g., HalfCheetah, CartPole), higher MVS corresponds to lower reward. Inverted pairs cluster near MVS $\approx 0$ regardless of reward loss.}
    \label{fig:combined}
\end{figure}

Phase~1 and Phase~2 independently establish two facts: MVS tracks violation severity in high-dimensional environments (Spearman $\rho$ up to $0.78$, 60 observations per pair), and those same violations degrade reward (13 of 16 pairs significant after FDR correction). Figure~\ref{fig:combined} puts the two together by plotting Phase~1 MVS against Phase~2 reward ratio for each condition.

The scatter reveals a clear pattern: in environments where MVS works correctly---primarily high-dimensional ones like HalfCheetah---points fan out to the right with increasing noise, and higher MVS corresponds to worse reward. In inverted environments, MVS stays near zero regardless of noise level, so reward degrades without a corresponding MVS signal. This is consistent with the inversion mechanism from Section~\ref{subsec:inversion}. The qualitative takeaway is that MVS is most informative in environments where the random forest cannot easily memorize the noise---precisely the higher-dimensional settings where diagnosis is most needed.


\subsection{Practical Utility: MVS-Guided Architecture Selection}
\label{subsec:utility}

The results so far show that MVS detects violations and that violations hurt reward. But can a practitioner actually \emph{use} the score? This is tested with a simple architecture-selection experiment: choose between a standard memoryless policy and one that receives a window of recent observations.

\paragraph{Setup.} CartPole-v1 is used under two observation conditions. In the \emph{full} condition the agent sees all four state variables (position, velocity, angle, angular velocity). In the \emph{masked} condition the two velocities are zeroed out, leaving only positions---a clean partial-observability setting where the current observation alone cannot determine the next state. For each condition, both a standard MLP policy and a history-augmented policy (current plus two prior observations concatenated) are trained. All runs use PPO, 100k steps, 5 seeds, 20 evaluation episodes.

\paragraph{MVS detection.} Table~\ref{tab:utility-mvs} shows MVS computed on trajectories from both random and trained policies.

\begin{table}[htbp]
    \centering
    \small
    \caption{\textbf{MVS detection of partial observability.} Masking velocities creates non-Markov observations. MVS is near zero for full observations and rises to $0.42$ under masking with a trained policy, correctly flagging the violation.}
    \label{tab:utility-mvs}
    \begin{tabular}{lcc}
        \toprule
        \textbf{Observation} & \textbf{Random Policy MVS} & \textbf{Trained Policy MVS} \\
        \midrule
        Full (4D)   & 0.000 & 0.000 \\
        Masked (positions only) & 0.002 & 0.421 \\
        \bottomrule
    \end{tabular}
\end{table}

MVS correctly flags the masked condition ($0.42$) and confirms that full observations are Markov ($0.00$). Interestingly, the signal is much stronger under a trained policy than a random one ($0.42$ vs.\ $0.002$), likely because a trained policy concentrates on a narrow region of state space where the missing velocity information matters more.

\paragraph{Architecture selection.} Table~\ref{tab:utility-reward} shows what happens when the MVS signal is acted upon.

\begin{table}[htbp]
    \centering
    \small
    \caption{\textbf{Architecture selection results.} Under full observations both policies hit 500. Under masking the standard policy collapses while the history-augmented policy fully recovers. PPO, 100k steps, 5 seeds, 20 evaluation episodes.}
    \label{tab:utility-reward}
    \begin{tabular}{llrr}
        \toprule
        \textbf{Observation} & \textbf{Policy} & \textbf{Mean Reward} & \textbf{Std} \\
        \midrule
        Full        & Standard           & 500.0 & 0.0 \\
        Full        & History-augmented   & 500.0 & 0.0 \\
        Masked      & Standard           & 43.1  & 1.2 \\
        Masked      & History-augmented   & 500.0 & 0.0 \\
        \bottomrule
    \end{tabular}
\end{table}

The takeaway is clear. When observations are Markov ($\mathrm{MVS} = 0$), a standard policy suffices and adding history buys nothing. When observations are non-Markov ($\mathrm{MVS} = 0.42$), the standard policy collapses to reward 43---a 91\% drop---while the history-augmented policy recovers fully to 500. A simple rule---use the standard architecture when MVS is near zero, switch to history augmentation otherwise---gets optimal performance in both cases without adding unnecessary complexity.

\section{Limitations and Future Directions}
\label{sec:limitations}

\paragraph{MVS inversion in low-dimensional environments.} The most significant limitation is the inversion phenomenon documented in Section~\ref{subsec:inversion}: in 8 of 16 environment--algorithm pairs, MVS \emph{decreases} as the true violation grows stronger. Because the random forest in Stage~1 is flexible enough to absorb the noise signal itself, the residuals end up cleaner than they should be---and Stage~2 finds nothing. This is not an implementation bug but a fundamental tension in any two-stage approach with a powerful first stage. Several directions seem worth exploring: restricting forest capacity (fewer trees, shallower depth) so it cannot latch onto temporal noise patterns; falling back to a linear first stage in low-dimensional settings where Markov dynamics are roughly linear anyway; ensembling across Stage~1 models of varying capacity; or collapsing to a single-stage direct comparison between Markov and history models, accepting higher false-positive rates in nonlinear-but-Markov systems.

\paragraph{AR(1) noise as the sole violation type.} MVS has been tested exclusively against AR(1) observation noise. AR(1) provides a clean, parameterized violation whose severity is controlled, but real-world non-Markov structure comes from many sources: sensor latency, frame stacking artifacts, communication delays, and genuinely missing state dimensions. Whether MVS generalizes to these settings is an open question. Frame stacking is especially interesting---it introduces a qualitatively different kind of history dependence that may be easier or harder to detect than smooth autocorrelation.

\paragraph{From diagnosis to remedy.} Section~\ref{subsec:utility} shows that MVS can guide architecture selection in a controlled setting, but fully closing the loop remains open. MVS could potentially run as an online diagnostic during training, triggering adaptive responses---switching to a history-augmented architecture when the score crosses a threshold, adjusting learning rates, or informing sensor suite design by comparing MVS across observation configurations. That said, the inversion problem must be resolved before MVS can serve as a reliable online signal in all environments.

\paragraph{Scalability.} The pipeline requires collecting a trajectory, computing random forest predictions, and fitting ridge regressions. All steps scale linearly in trajectory length and polynomially in observation dimensionality. For the environments tested (up to 17 dimensions), computation is negligible next to RL training time. Scaling to image observations would require a representation learning step before applying MVS, which introduces its own assumptions and potential failure modes.

\paragraph{Algorithm coverage.} PPO, A2C, and SAC---representative on-policy and off-policy methods---are evaluated here. How Markov violations interact with model-based RL, offline RL, or multi-agent settings remains unexplored. Model-based methods are particularly interesting: their explicit dynamics models could either amplify non-Markov noise or partially compensate for it.

\paragraph{Cross-validation and noise design.} Stage~2 selects ridge regularization via leave-one-out cross-validation. Because adjacent time points share lagged features, LOO can leak temporal information; blocked or rolling CV would provide a stricter protocol. Additionally, AR(1) noise with larger $\alpha$ has higher marginal variance, so Phase~2 reward degradation conflates temporal correlation with noise power. A design matching marginal variance across $\alpha$ while varying only autocorrelation would isolate the non-Markov contribution.

\paragraph{Theoretical guarantees.} MVS currently lacks formal statistical guarantees. The two-stage procedure creates dependencies between the residuals and the ridge models that complicate standard hypothesis testing. Establishing Type~I and Type~II error rates, or connecting MVS to conditional independence tests with known power properties, would put the method on firmer theoretical ground.

\section{Conclusion}
\label{sec:conclusion}

A prediction-based Markov Violation Score is introduced that quantifies non-Markovian structure in RL observation trajectories. The two-stage approach---random forest residualization followed by ridge regression comparison---yields a bounded, interpretable scalar: zero when the process is Markov, increasing with the severity of history dependence.

Across six environments, three algorithms, and controlled AR(1) noise, MVS successfully detects violations in 7 of 16 environment--algorithm pairs, with Spearman correlations up to $\rho = 0.78$ between noise intensity and the score. In these environments, higher MVS corresponds to worse policy performance. Phase~2 results confirm that these violations have real consequences: 13 of 16 pairs show statistically significant reward degradation under noise. In a practical utility experiment, MVS is shown to guide architecture selection---when partial observability renders observations non-Markovian, the score flags the problem and a history-augmented policy chosen accordingly recovers the lost performance entirely. At the same time, an inversion phenomenon in low-dimensional environments limits detection in nearly half the tested conditions, pointing to a fundamental tension between flexible first-stage modeling and preserving the signal that the second stage needs.

The candid analysis of when MVS works and when it does not is a contribution in its own right. Reliable detection of history dependence is a prerequisite for principled mitigation, and understanding why prediction-based approaches fail in certain regimes should inform the design of more robust diagnostics going forward.

\bibliography{references}
\bibliographystyle{rlj}

\beginSupplementaryMaterials

\section{Declaration of LLM Usage}
Large language models (GPT-4, Claude) were used during manuscript preparation for grammar correction and revising passive voice constructions. All scientific content, experimental design, implementation, and analysis are the sole work of the author.

\section{Implementation and Reproducibility Details}
All RL agents are trained using Stable-Baselines3~\citep{raffin_stable-baselines3_2021} with default hyperparameters for each algorithm (PPO, A2C, SAC). The AR(1) noise wrapper, MVS computation pipeline, and analysis scripts are included in the supplementary source code. Random seeds are fixed for reproducibility; each condition is evaluated over 10 independent seeds.

\end{document}